\documentclass[letterpaper, 10 pt, conference]{ieeeconf}
\IEEEoverridecommandlockouts
\usepackage{cite}
\usepackage{amsmath,amssymb,amsfonts}
\usepackage{graphicx}
\usepackage{algorithmic}
\usepackage{textcomp}
\usepackage{xcolor}
\usepackage{textgreek}
\usepackage{tikz}
\usetikzlibrary{shapes.geometric, arrows.meta, positioning}
\usepackage{booktabs}
\usepackage{multirow}
\usepackage[normalem]{ulem}

\begin{document}

\title{\LARGE \bf
Do Open-Vocabulary Detectors Transfer to Aerial Imagery? A Comparative Evaluation
}

\author{Christos Tsourveloudis\\[1ex]
\small School of Electrical and Computer Engineering, National Technical University of Athens, Greece\\
\small el21059@mail.ntua.gr
}

\maketitle

\begin{abstract}
Open-vocabulary object detection (OVD) enables zero-shot recognition of novel categories through vision-language models, achieving strong performance on natural images. However, transferability to aerial imagery remains unexplored. We present the first systematic benchmark evaluating five state-of-the-art OVD models on the LAE-80C aerial dataset (3,592 images, 80 categories) under strict zero-shot conditions. Our experimental protocol isolates semantic confusion from visual localization through Global, Oracle, and Single-Category inference modes. Results reveal severe domain transfer failure: the best model (OWLv2) achieves only 27.6\% F1-score with 69\% false positive rate. Critically, reducing vocabulary size from 80 to 3.2 classes yields 15× improvement, demonstrating that semantic confusion is the primary bottleneck. Prompt engineering strategies such as domain-specific prefixing and synonym expansion, fail to provide meaningful performance gains. Performance varies dramatically across datasets (F1: 0.53 on DIOR, 0.12 on FAIR1M), exposing brittleness to imaging conditions. These findings establish baseline expectations and highlight the need for domain-adaptive approaches in aerial OVD.

\end{abstract}

\section{Introduction}
\label{sec:intro}

Unmanned Aerial Vehicles (UAVs) and satellite platforms are increasingly deployed for time-critical applications such as disaster response, infrastructure monitoring, and search-and-rescue operations. In these scenarios, robust object detection is essential—yet the objects of interest are often unpredictable. A disaster response team may need to locate "overturned vehicles" or "temporary shelters" that were never part of the detector's training vocabulary. Traditional closed-set detectors, which recognize only predefined categories, fail catastrophically when confronted with such novel objects, limiting their utility in dynamic real-world deployments.~\cite{leng2024recent}.
At the same time, collecting and annotating large, diverse datasets to cover every possible category is expensive, time-consuming, and often impractical.

These constraints have motivated the development of Open-Vocabulary Object Detection (OVD). Instead of learning only from labeled training categories, OVD methods incorporate semantic knowledge from vision-language models (VLMs) to relate visual features to natural language descriptions~\cite{wu2024towards}. This allows them to detect objects defined solely by text prompts, enabling zero-shot recognition of categories not explicitly annotated during training~\cite{zareian2021openvoc}. By leveraging linguistic understanding, OVD systems offer a more scalable and flexible alternative that better reflects the complexity of the real world.

However, a critical question remains unanswered: Do these OVD models, trained predominantly on ground-level imagery, transfer effectively to the aerial domain? Aerial imagery presents distinct challenges, including extreme scale variation, strict nadir viewpoints, and dense object clustering—conditions~\cite{Ding2022challenge}. Moreover, semantic categories in aerial datasets often rely on fine-grained visual attributes or domain-specific terminology that may not align with the natural language priors encoded in VLMs. 

Despite the rapid growth of OVD methods, their zero-shot transferability to aerial platforms remains largely unexplored. While recent work has proposed aerial-specific OVD architectures trained on remote sensing data, no study has systematically evaluated whether general-purpose VLM-based detectors can provide a viable baseline for UAV applications, deployable without any aerial-specific fine-tuning.

To better understand the feasibility of language-driven detection in these scenarios, this work investigates how existing vision–language models perform when applied directly to UAV imagery, without additional training. We also explore how different prompting strategies influence results, aiming to establish an initial reference point for applying open-vocabulary object detection techniques in aerial applications.

The remainder of this paper is organized as follows: Section~\ref{sec:related_work} reviews prior research Open-Vocabulary Object Detection. Section~\ref{sec:methodology} details our methodology, including the taxonomic challenges of the LAE-80C benchmark, the architectures of the evaluated models, and our zero-shot experimental design. Section~\ref{sec:results} presents a quantitative and qualitative analysis of model performance, focusing on the precision-recall trade-offs and the impact of prompting strategies. Section V discusses the implications of our findings for real-world UAV deployment, and Section VI concludes the paper with directions for future work.

\section{Related Work}
\label{sec:related_work}

The emergence of large-scale vision-language models (VLMs) such as CLIP~\cite{radford2021clip} and ALIGN~\cite{jia2021scaling} has fundamentally transformed object detection. By pretraining on hundreds of millions of image-text pairs, these models learn a shared embedding space where visual regions and natural language descriptions are semantically aligned. This enables a new detection paradigm: rather than training classifiers for fixed categories, detectors can match image regions to arbitrary text prompts at inference time.

OVD methods generally fall into two categories: (i) two-stage approaches, such as ViLD, which use knowledge distillation from powerful VLMs but incur substantial computational cost~\cite{gu2022vild}; and (ii) end-to-end frameworks, such as GLIP~\cite{li2022glip}, Grounding DINO~\cite{liu2024groundingdino}, YOLO-World~\cite{cheng2024yolow} and YOLOE~\cite{cheng2025yoloe} which integrate language grounding directly into the detection pipeline for streamlined inference. Recent studies also underscore the role of linguistic clarity; for instance, Hosoya et al.~\cite{hosoya2024openvoc} show that OVD can underperform on ambiguous categories lacking precise descriptions—a key issue in remote sensing, where many targets defy specific naming.

In the aerial domain, Teacher–student frameworks like CastDet employ VLMs pretrained on remote-sensing image–text pairs to create pseudo-labels for new aerial categories, facilitating detection without closed-set annotations~\cite{li2024toward}. Complementary techniques focus on strengthening semantic alignment through description regularization and similarity-aware losses, as in DescReg~\cite{zang2024zero}. Specialized architectures such as OVA-DETR~\cite{wei2024ova} and UAV-OVD~\cite{tao2025uav} incorporate region–text contrastive learning and multi-level fusion to enhance small object detection while preserving real-time efficiency. Other efforts address cross-view mismatches by aligning ground- and aerial-view embeddings~\cite{kini2025cross}. With regard to large foundation multimodal models, they have been tested for UAV tasks~\cite{hannan2025foundation} but often lag in localizing small objects compared to tailored OVD detectors for aerial data.

Progress in open-vocabulary aerial perception also relies on robust data resources. Traditional benchmarks such as DOTA~\cite{xia2018dota}, DIOR~\cite{li2020dior}, VisDrone~\cite{zhu2020visdrone}, FAIR1M~\cite{sun2022fair1m} and xView~\cite{lam2018xview} were originally built for closed-set detection but have since been augmented with language annotations. More recent pipelines, such as LAE-1M~\cite{pan2025locate} and OS-W2S~\cite{wei2025osw2s}, automate text-aligned label generation using VLMs, enabling scalable supervision in remote sensing.

The OVD paradigm can also be extended beyond bounding box detection to tasks such as segmentation and tracking. AeriaICLIP supports lightweight open-vocabulary segmentation tailored for UAV hardware~\cite{jia2025aeriaiclip}, while Follow Anything (FAn) combines open-set detection with tracking for dynamic navigation~\cite{maalouf2024follow}. 

In summary, the field has advanced through language-aligned learning, adaptive architectures and automated data pipelines. Nonetheless, persistent hurdles remain, including unreliable zero-shot performance due to vague aerial category descriptions, the difficulty of managing dense small objects in high-altitude views, and strict UAV constraints on computation and power. 

\section{Methodology}
\label{sec:methodology}

\subsection{Dataset: LAE-80C}

We evaluate zero-shot open-vocabulary detection on the LAE-80C benchmark, the evaluation subset of the Locate Anything on Earth (LAE-1M) dataset. LAE-80C aggregates validation splits from four established aerial object detection datasets: DOTA-v2.0, DIOR,FAIR1M and xView, resulting in 3,592 images with 86,558 annotated object instances across 80 categories~\cite{pan2025locate}.

We specifically selected LAE-80C for this study because it represents a rigorous stress test for zero-shot generalization. By aggregating data from diverse sources, it introduces high variance in image quality, sensor look-angle, and ground sampling distance—spanning from low-altitude drone imagery to satellite data. This ensures that our evaluation measures genuine model robustness across domains, rather than performance on a single, homogeneous dataset (like DIOR) which current state-of-the-art models may have already saturated~\cite{Chen2023MDCT}.

While the benchmark aims to provide a comprehensive label space, our analysis of the 80 classes reveals significant taxonomic complexities that pose unique challenges for Vision-Language alignment:

\begin{itemize}
    \item \textbf{Hierarchical Overlap:} The label space contains both superclass and subclass definitions. For instance, the generic class \texttt{vehicle} co-exists with specific classes such as \texttt{passenger car}, \texttt{small car}, and \texttt{pickup truck}. This forces the model to perform fine-grained disambiguation in a zero-shot setting, often leading to semantic confusion.
    
    \item \textbf{Attribute Dependence:} Several categories are distinguished solely by transient states rather than structural differences. Notably, \texttt{working chimney} and \texttt{unworking chimney} are treated as distinct classes, requiring the detector to identify the presence of smoke (an attribute) rather than simply localizing the object.
    
    \item \textbf{Fine-Grained Clusters:} The dataset includes dense clusters of visually identical categories, particularly in the maritime domain (e.g., distinguishing between \texttt{liquid cargo ship}, \texttt{oil tanker}, and \texttt{dry cargo ship}).
\end{itemize}

These characteristics make LAE-80C particularly challenging for zero-shot OVD, as models must disambiguate semantically related categories without the benefit of task-specific training.

\subsection{Models Evaluated}

We evaluate five representative open-vocabulary detectors spanning different architectural paradigms and efficiency trade-offs (Table~\ref{tab:model_comparison}). All models were pretrained exclusively on natural, ground-level images and evaluated in a strict zero-shot setting—no fine-tuning on LAE-80C or any remote sensing data was performed.

\textbf{Grounding DINO} employs a Transformer-based architecture with a Swin-L backbone (218M parameters) and introduces deep cross-modality fusion via a language-guided query selection mechanism. Text and image features interact at multiple decoder layers through cross-attention, enabling fine-grained semantic grounding. While Grounding DINO achieves state-of-the-art performance on COCO (mAP 52.5 with zero-shot COCO class names), its computational cost limits real-time applicability~\cite{liu2024groundingdino}.
    
\textbf{OWLv2} formulates detection as dense image-text matching using a ViT-L/14 backbone (428M parameters). Unlike region-proposal methods, OWLv2 processes image patches and text tokens jointly through a contrastive objective trained on the WebLI dataset (10B image-text pairs). This massive-scale pretraining enables robust handling of long-tail categories~\cite{minderer2023scaling}.
    
\textbf{YOLO-World} and \textbf{YOLOE-11} represent the real-time detection family. YOLO-World is built on YOLOv8 and introduces RepVL-PAN, which enables "offline vocabulary encoding"—text embeddings are precomputed and merged into convolutional weights during inference~\cite{cheng2024yolow}. YOLOE, based on the YOLOv11-L backbone, advances this further by integrating the efficiency of the YOLOv11 architecture with a unified "Seeing Anything" prompt mechanism~\cite{cheng2025yoloe}. This allows it to handle text, visual, and prompt-free inputs with significantly fewer parameters than its predecessors while maintaining high inference speeds on consumer GPUs.

\textbf{LLMDet} integrates a frozen Large Language Model to enhance semantic reasoning. The architecture builds upon Grounding DINO, utilizing a \textbf{Swin-L} visual encoder to extract region features, while leveraging an LLM to generate detailed image captions and enriched region descriptions~\cite{fu2025llmdet}. This approach aims to leverage the LLM's linguistic understanding to resolve ambiguous category definitions, which are particularly relevant for LAE-80C's attribute-dependent classes.

A critical commonality across all selected models is their training domain. All were pre-trained primarily on natural, ground-level imagery. Consequently, they share a distinct domain gap when applied to aerial imagery, where objects appear strictly from a top-down nadir perspective and lack the canonical side-profile features (e.g., wheels on a car) that these models rely on for feature extraction. Our objective is to evaluate the zero-shot transferability of general-purpose foundation models. We aim to determine if the 'Open Vocabulary' promise of mainstream computer vision holds up in the aerial domain, or if the domain gap is substantial enough to make these 'universal' detectors unusable without specialized retraining.

\begin{table}[t]
\centering
\caption{Model Specifications}
\label{tab:model_comparison}
\small
\begin{tabular}{lcccc}
\toprule
\textbf{Model} & \textbf{Backbone} & \textbf{Params} & \textbf{FPS} \\
\midrule
Grounding DINO & Swin-L & 218M & 3.2 \\
OWLv2 & ViT-L/14 & 428M & 8.5 \\
YOLO-World & CSPDarknet & 60M & 52.1 \\
YOLO-E & C3k2 & 26M & 65.3 \\
LLMDet & Swin-L & 435M & 3.4 \\
\bottomrule
\end{tabular}
\end{table}

\subsection{Experimental Design}

To systematically evaluate the zero-shot capabilities of the selected models, we design three evaluation modes that progressively reduce semantic complexity:

\begin{enumerate}
    \item \textbf{Global Inference:} The model is prompted with the full vocabulary of 80 category names simultaneously for every image. This represents the standard zero-shot scenario where the detector must filter true positives from 79 negative classes per ground-truth category, testing both localization and classification under maximum semantic confusion.
    
    \item \textbf{Oracle Inference:} The model is prompted only with the ground-truth categories present in the specific image. By removing "distractor" classes, this mode isolates the model's visual localization capability from the confusion caused by the dataset's overlapping taxonomy.
    
    \item \textbf{Single-Category Oracle Inference:} The model performs inference iteratively, prompting one class at a time for all classes present in a given image. This computationally intensive method tests whether the model's attention mechanism degrades when processing long text sequences versus focused single-token prompts.
\end{enumerate}

Standard zero-shot detection relies on fixed class names provided by the dataset taxonomy. However, this rigid approach often fails due to two distinct gaps: the \textit{domain gap} (natural vs. aerial imagery) and the \textit{lexical gap} (mismatched vocabulary between the model's training data and the dataset labels). To mitigate these issues, we implemented a two-stage prompt engineering strategy:

To align the text embeddings with the overhead visual features characteristic of the LAE-80C dataset, we applied template-based prompt engineering. Beyond raw class names, we evaluated the impact of appending domain-specific prefixes, specifically transforming prompts to "\textit{Aerial view of \{category\}}". This contextual cue aims to bias the model towards top-down features rather than the side-profile features typical of its ground-level pre-training data.

To address the \textit{lexical rigidity} of the LAE-80C taxonomy—where a model might fail to detect a "Ship" simply because it was prompted with the word "Vessel"—we implemented a synonym expansion protocol. We constructed a mapping dictionary, which associates each of the 80 class IDs with a list of semantic equivalents. 

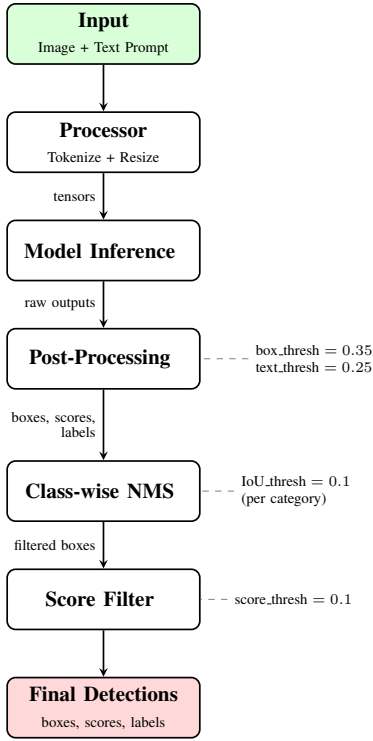
\begin{figure}[!h]
\centering
\resizebox{0.6\columnwidth}{!}{
\begin{tikzpicture}[
    >=stealth,
    auto,
    node distance=1.8cm,
    block/.style={draw, thick, rectangle, rounded corners, 
                  align=center, minimum width=3.2cm, minimum height=1.0cm},
    line/.style={draw, thick, ->},
    dashedblock/.style={draw, thick, dashed, rectangle, 
                  align=center, inner sep=0.2cm, rounded corners},
    paramblock/.style={draw=none, align=left, font=\scriptsize}
]

\node[block, fill=green!15] (input) {
  \textbf{Input}\\
  \scriptsize Image + Text Prompt
};

\node[block, below of=input] (preprocess) {
  \textbf{Processor}\\
  \scriptsize Tokenize + Resize
};

\node[block, below of=preprocess] (inference) {
  \textbf{Model Inference}
};

\node[block, below of=inference] (postprocess) {
  \textbf{Post-Processing}
};

\node[paramblock, right of=postprocess, node distance=3.5cm] (postparams) {
  $\text{box\_thresh} = 0.35$\\
  $\text{text\_thresh} = 0.25$
};

\node[block, below of=postprocess, node distance=2.2cm] (nms) {
  \textbf{Class-wise NMS}
};

\node[paramblock, right of=nms, node distance=3.2cm] (nmsparams) {
  $\text{IoU\_thresh} = 0.1$\\
  (per category)
};

\node[block, below of=nms, node distance=1.8cm] (filter) {
  \textbf{Score Filter}
};

\node[paramblock, right of=filter, node distance=3.2cm] (filterparams) {
  $\text{score\_thresh} = 0.1$
};

\node[block, below of=filter, fill=red!15] (output) {
  \textbf{Final Detections}\\
  \scriptsize boxes, scores, labels
};

\draw[line] (input) -- (preprocess);

\draw[line] (preprocess) -- 
  node[left]{\scriptsize tensors} (inference);

\draw[line] (inference) -- 
  node[left]{\scriptsize raw outputs} (postprocess);

\draw[line] (postprocess) -- 
  node[left, pos=0.4]{\scriptsize boxes, scores,} 
  node[left, pos=0.6]{\scriptsize labels} (nms);

\draw[line] (nms) -- 
  node[left]{\scriptsize filtered boxes} (filter);

\draw[line] (filter) -- (output);

\draw[dashed, gray] (postparams.west) -- (postprocess.east);
\draw[dashed, gray] (nmsparams.west) -- (nms.east);
\draw[dashed, gray] (filterparams.west) -- (filter.east);

\end{tikzpicture}
}
\caption{Detection pipeline for open-vocabulary aerial object detection. The pipeline progressively filters detections using box confidence, text alignment, class-wise NMS, and final score thresholds.}
\label{fig:detection_pipeline}
\end{figure}

The complete detection and evaluation pipeline is illustrated in Figure~\ref{fig:detection_pipeline}. Each image and its corresponding text prompts are first tokenized and resized by the model-specific processor. Following inference, predictions undergo a multi-stage filtering process: initial post-processing applies box confidence ($\text{box\_thresh} = 0.35$) and text-alignment thresholds ($\text{text\_thresh} = 0.25$), followed by class-wise Non-Maximum Suppression (NMS) with $\text{IoU} = 0.1$ to eliminate duplicate detections within each category. Finally, a score threshold of $0.1$ filters low-confidence predictions before evaluation. Evaluation was performed using standard detection metrics including Precision, Recall, and F1-score. 

\textbf{Intersection over Area (IoA):} Unlike standard COCO evaluation which strictly uses Intersection over Union (IoU), we utilized Intersection over Area (IoA) with a threshold of $0.7$. We selected IoA over IoU to account for the significant scale variance in satellite imagery, where small objects (like vehicles) may be partially occluded or densely clustered. In such cases, IoA provides a more robust measure of whether a prediction effectively "covers" the ground truth object without penalizing minor misalignment.

\textbf{Confidence Threshold:} Detections were filtered with a confidence score threshold of $\mathbf{0.1}$. This relatively low threshold was chosen to capture the full range of the model's recall capabilities before applying stricter filtering.

\section{Results}
\label{sec:results}

In this section, we present the quantitative evaluation of state-of-the-art open-vocabulary object detectors on the aerial test set. We assess performance using standard metrics: Precision, Recall, and F1-score, alongside a granular analysis of True Positives (TP), False Positives (FP), and False Negatives (FN) to understand specific failure modes.

\subsection{Overall Performance and Domain Gap}

\begin{figure}[!h]
    \centering
    \includegraphics[width=\columnwidth]{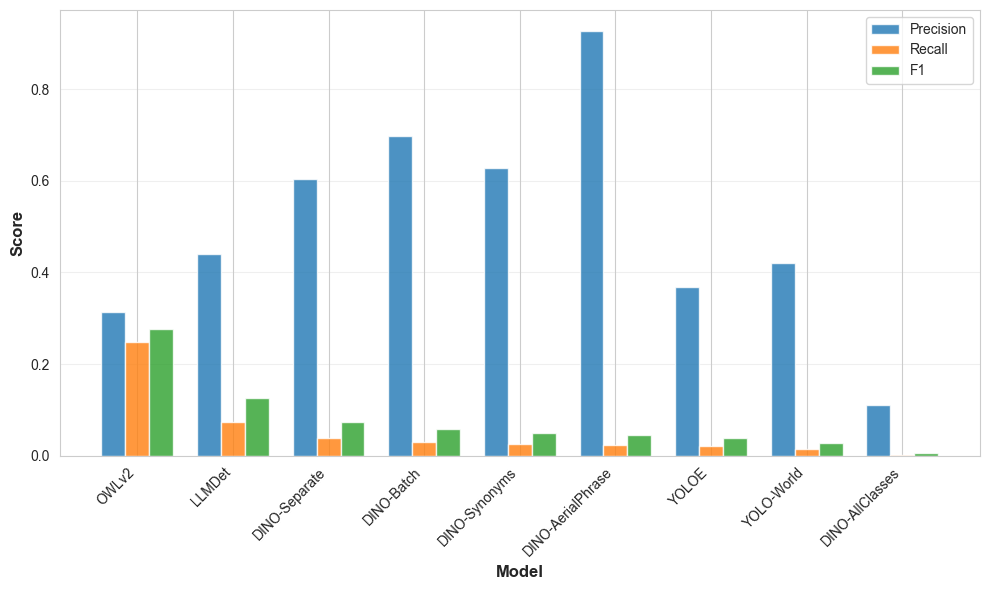}
    \caption{Comparison of Precision, Recall, and F1 Score for all evaluated models.}
    \label{fig:f1_pr}
\end{figure}

\begin{figure}[!h]
    \centering
    \includegraphics[width=\columnwidth]{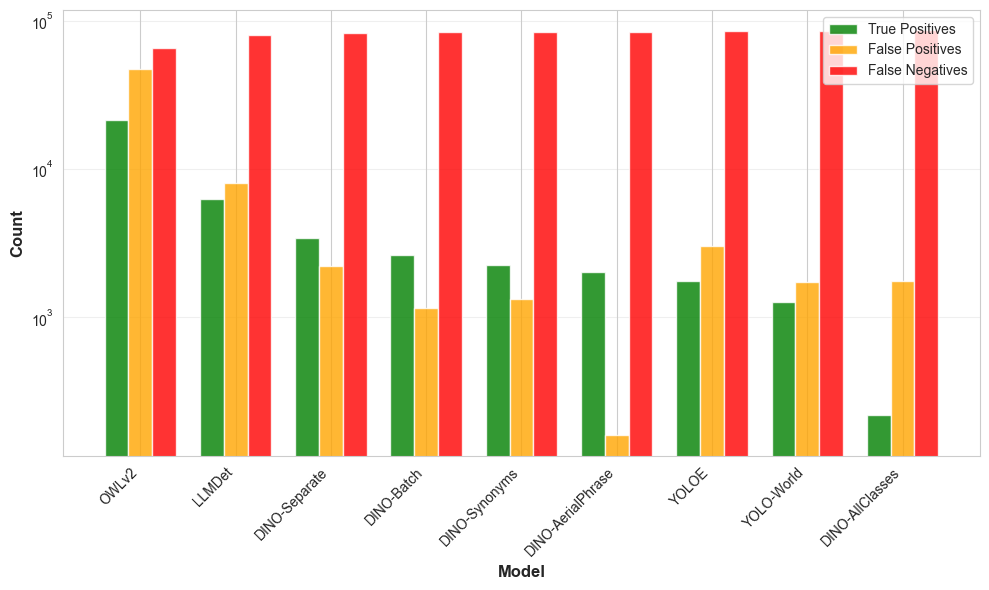}
    \caption{True Positive (TP), False Positive (FP), and False Negative (FN) counts across models.}
    \label{fig:tp_fp_fn}
\end{figure}

Table~\ref{tab:main_results} presents the zero-shot performance of all evaluated models on LAE-80C. Figures~\ref{fig:f1_pr} and~\ref{fig:tp_fp_fn} provide a visual summary of the results in Table~\ref{tab:main_results}. The results reveal a severe domain transfer gap: all models exhibit dramatically degraded performance compared to their reported natural-image benchmarks. Despite the overall poor performance, some useful insights emerge.

\begin{table}[!h]
    \centering
    \caption{Zero-shot performance of the evaluated models.}
    \label{tab:main_results}
    \resizebox{\columnwidth}{!}{%
    \begin{tabular}{lcccccc}
        \toprule
        \textbf{Model} & \textbf{Precision} & \textbf{Recall} & \textbf{F1 Score} & \textbf{TP} & \textbf{FP} & \textbf{FN} \\
        \midrule
        OWLv2 & 0.313 & \textbf{0.247} & \textbf{0.276} & \textbf{21,408} & 47,058 & \textbf{65,150} \\
        LLMDet & 0.441 & 0.073 & 0.125 & 6,308 & 8,009 & 80,250 \\
        DINO-Separate & 0.604 & 0.039 & 0.074 & 3,409 & 2,233 & 83,149 \\
        DINO-Batch & 0.697 & 0.031 & 0.059 & 2,650 & 1,151 & 83,908 \\
        DINO-Synonyms & 0.628 & 0.026 & 0.050 & 2,244 & 1,327 & 84,314 \\
        DINO-AerialPhrase & \textbf{0.926} & 0.023 & 0.045 & 2,014 & \textbf{160} & 84,544 \\
        YOLOE & 0.367 & 0.020 & 0.039 & 1,767 & 3,045 & 84,791 \\
        YOLO-World & 0.421 & 0.015 & 0.028 & 1,269 & 1,744 & 85,289 \\
        DINO-AllClasses & 0.111 & 0.003 & 0.005 & 218 & 1,748 & 86,340 \\
        \bottomrule
    \end{tabular}%
    }
\end{table}

\textbf{OWLv2} emerges as the most effective model for aerial zero-shot detection, achieving F1=27.6\% and recall=24.7\%—more than 3× higher recall than any other model. Its architecture, based on dense image-text matching across all image patches, appears more robust to the distributional shift than region-proposal methods. However, this comes at a severe cost: 47,058 false positives against 21,408 true positives (69\% false positive rate), indicating the model struggles to distinguish aerial objects from textured backgrounds. This high-recall, low-precision regime may be acceptable for human-in-the-loop workflows where operators can filter false alarms, but cannot be used as a fully autonomous deployment.

\textbf{LLMDet} achieves the second-best F1-score (12.5\%), offering a more balanced precision-recall trade-off (P=44.1\%, R=7.3\%) than OWLv2. The integration of LLMs for semantic reasoning provides improvements over baseline CLIP embeddings, though the model still fails to detect over 92\% of ground-truth objects. The LLM component appears to help with disambiguation (higher precision than YOLO models) but cannot overcome the fundamental visual domain gap.

\textbf{YOLO Family} (YOLO-World, YOLO-E) exhibits near-catastrophic failure in zero-shot aerial transfer, with F1-scores below 4\% despite competitive performance on natural-image benchmarks. YOLO-E's improvements over YOLO-World (F1: 3.9\% vs 2.8) are marginal, suggesting the bottleneck is architectural rather than  in the training methodology.

\textbf{Grounding DINO} performance varies dramatically based on inference mode and prompting strategy (analyzed in detail in Subsection C). In Global mode (all 80 classes), the model suffers catastrophic failure, detecting only 218 out of 86,558 objects (0.25\% recall, F1=0.5\%). However, in Oracle mode with ground-truth classes only, performance improves to F1=5.9-7.4\%—still poor in absolute terms, but representing a 15× improvement over Global mode. This dramatic variance reveals that semantic confusion, rather than visual feature extraction failure, is the dominant bottleneck for Transformer-based detectors.

\subsection{Precision-Recall Trade-off}

\begin{figure}[!h]
    \centering
    \includegraphics[width=\columnwidth]{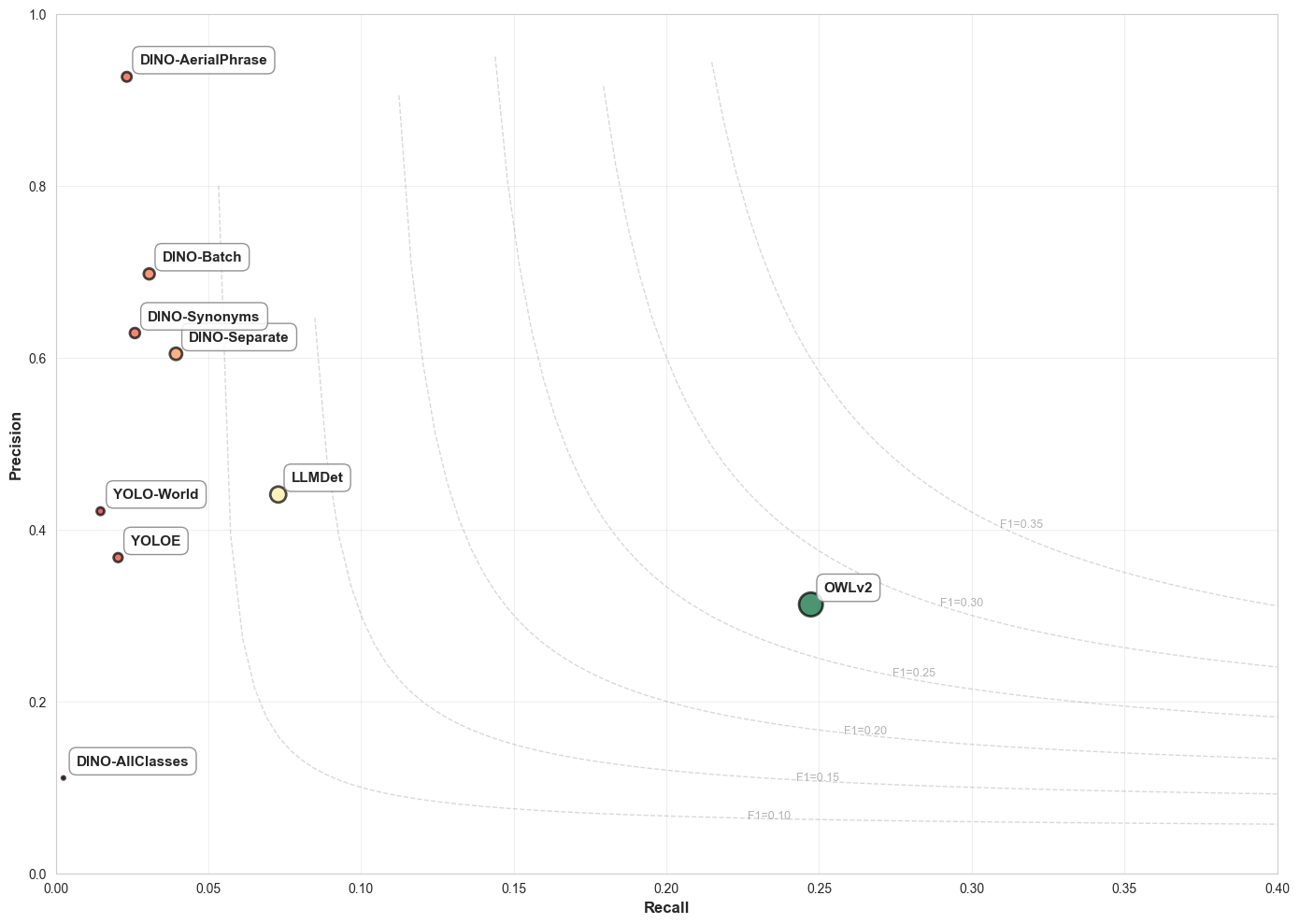}
    \caption{Precision-Recall Trade-off. DINO variants optimize for precision at the expense of coverage, while OWLv2 sacrifices precision for higher recall. Ideally, a model would appear in the top-right corner.}
    \label{fig:pr_scatter}
\end{figure}

The trade-off between detection sensitivity and correctness is visualized in Figure~\ref{fig:pr_scatter}. 

The results reveal two distinct clusters of model behavior:

\textbf{The "Conservative" Cluster (Grounding DINO, YOLO-World, YOLO-E):} These models prioritize high precision (36-93\%) at catastrophic recall cost (0.3-3.9\%). The most extreme example is Grounding DINO with aerial prefix prompting, which achieves 92.6\% precision by detecting only 2.3\% of objects (2,014 TP vs 84,544 FN). While this conservatism minimizes false alarms (only 160 FP), it renders the model insufficient for real-world coverage requirements. This behavior suggests the models emit predictions only when visual-semantic alignment exceeds an extremely high confidence threshold—a threshold rarely met in domain-shifted aerial imagery.

\textbf{The "Aggressive" Cluster (OWLv2):} OWLv2 occupies the opposite extreme, prioritizing coverage at the cost of precision. With 24.7\% recall, it detects more objects than all other models combined, but pays the price with 47,058 false positives. For every 1 true detection, OWLv2 generates 2.2 false alarms. For certain aerial applications like disaster response \& search-and-rescue, this regime may be preferable, as missing critical objects carries higher cost than false alarms that humans can filter.

LLMDet occupies an intermediate position (P=44.1\%, R=7.3\%), but closer to the conservative cluster. Notably, no model achieves F1-score above 28\%, and the precision-recall frontier is severely degraded compared to natural-image performance. 

\begin{figure}[!h]
    \centering
    \includegraphics[width=0.69\columnwidth]{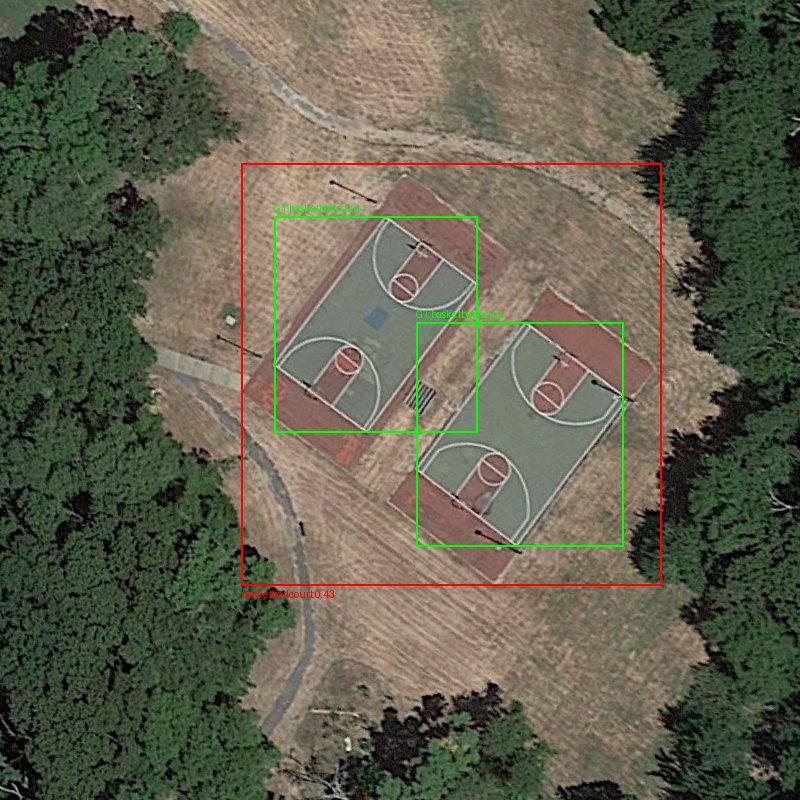}
    \caption{
    Conservative detection behavior. Grounding DINO collapses multiple nearby instances into a single high-confidence detection, prioritizing precision over instance-level recall.
}
    \label{fig:conservative_bboxes}
\end{figure}

\begin{figure}[!h]
    \centering
    \includegraphics[width=0.69\columnwidth]{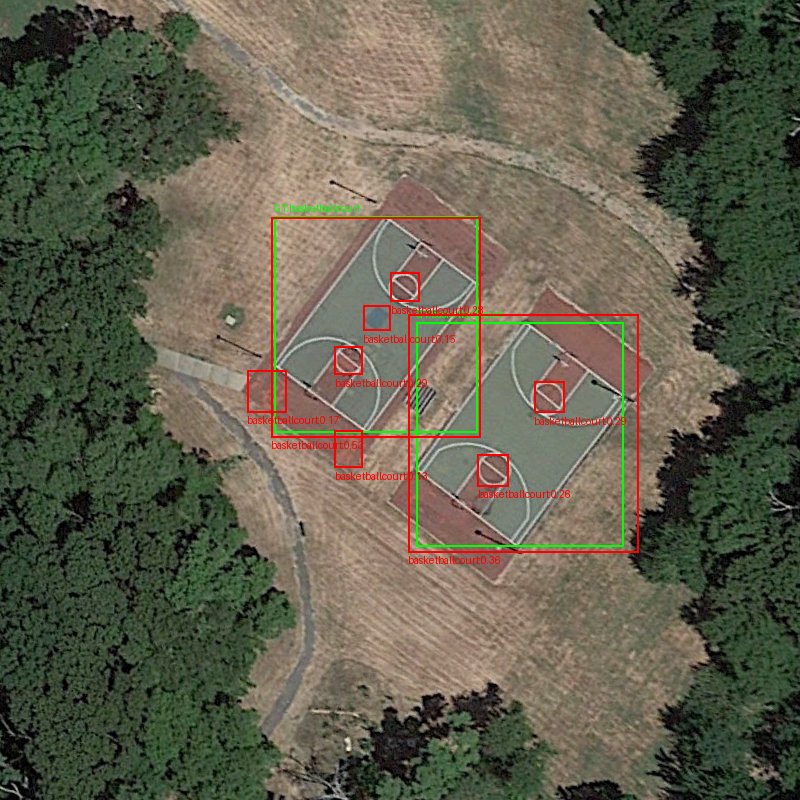}
    \caption{
    Aggressive detection behavior. OWLv2 separates individual instances but produces numerous false positives, illustrating the high-recall, low-precision regime.
}
    \label{fig:aggressive_bboxes}
\end{figure}

The ideal operating point (high precision and high recall, upper-right quadrant) remains completely unreachable for all evaluated architectures. This trade-off is further illustrated qualitatively in Figures~\ref{fig:conservative_bboxes} and~\ref{fig:aggressive_bboxes}. Conservative models (Grounding DINO) collapse multiple nearby instances into a single high-confidence detection, producing one large bounding box covering both basketball courts, whereas the aggressive OWLv2 correctly separates the two distinct courts but generates many additional false positives in the process.

\subsection{Impact of Inference Modes and Prompting Strategies}
\label{subsec:ablations}

We conduct systematic ablations using Grounding DINO to isolate the impact of vocabulary size and prompt engineering. From the results of Table~\ref{tab:main_results} we can exctract the following results:

\subsubsection{\textbf{Global vs Oracle Inference}}
The most striking finding is the catastrophic effect of vocabulary size. Reducing the prompt from 80 classes (Global mode) to an average of 3.2 ground-truth classes per image (Oracle mode) produces a 15× improvement in F1-score (0.5\% → 7.4\%). This massive gap confirms that semantic confusion, not visual localization failure, is the primary bottleneck for open-vocabulary detection in aerial imagery. When presented with all 80 LAE-80C categories simultaneously, Grounding DINO's cross-modal attention mechanism breaks down. The 80-token text sequence diffuses attention across semantically similar categories, preventing confident predictions even when visual features are correctly extracted. In Oracle mode, with only 3-4 highly relevant categories, the model can focus its limited attention budget on meaningful distinctions.

\subsubsection{\textbf{Single-Category Oracle vs Batched Oracle}}
Processing each ground-truth class in a separate forward pass (Single Oracle) outperforms batching all ground-truth classes together (F1: 7.4\% vs 5.9\%). This 1.5 percentage point gap suggests that even 3-4 simultaneous prompts introduce mild attention competition within the Transformer's cross-attention layers. However, Single Oracle incurs higher computational cost (one forward pass per class), making batched Oracle more practical for real-world deployment despite the slight performance penalty.

\subsubsection{\textbf{Synonym Expansion (DINO-Synonyms)}}
Expanding prompts with semantic equivalents degrades performance (F1: 5.9\% → 5.0\%, recall: 3.1\% → 2.6\%). Analysis reveals two failure modes:

Many synonyms might introduce unintended meanings. For example, expanding "ship" to include "boat" triggers false positives on small recreational watercraft that aren't part of LAE-80C's maritime taxonomy (which focuses on commercial vessels). The synonym "vessel" might match with industrial storage tanks in certain contexts.

When multiple synonyms are processed simultaneously (e.g., "airplane, aircraft, plane, jet"), the model must distribute confidence across four text embeddings rather than concentrating on one. This appears to reduce per-synonym activation strength, overall recall.

\subsubsection{\textbf{"Aerial view of" Prefix (DINO-AerialPhrase)}}
Prepending "Aerial view of" to class names produces a paradoxical result: precision increases dramatically (60.4\% → 92.6\%) while recall decreases (3.9\% → 2.3\%), resulting in lower F1-score (7.4\% → 4.5\%). This counter-intuitive outcome reveals a fundamental limitation of text-based domain adaptation:
The phrase "Aerial view of" seems to over-constrain the model's matching criteria. By explicitly biasing toward "aerial" semantics, it filters out predictions with any ground-level visual characteristics—even when those features are present and useful in overhead imagery. Essentially, the model becomes hyper-conservative, emitting predictions only when it detects both the object and an explicit top-down view.
Additionally, the phrase "aerial view" is likely under-represented in CLIP's training corpus, which mostly contains ground-level photos with captions like "a photo of X" rather than "aerial view of X". This causes the text encoder to produce weaker, less informative embeddings, further degrading matching quality.
This suggests that general-purpose VLMs cannot be steered toward aerial semantics through text prompts alone because their visual encoders are pretrained exclusively on ground-level features.

Prompt engineering provides minimal gains and often harms performance when domain-specific terms misalign with the model's pretraining distribution. The dominant factor is vocabulary size: reducing semantic confusion from 80 overlapping classes to 3-4 relevant classes yields a 15× improvement, completely dwarfing any prompting strategy effect. This suggests that future aerial OVD systems should focus on intelligent vocabulary filtering (e.g., scene-specific class subsets based on context) rather than prompt template engineering.

\subsection{Sensitivity to Dataset Distribution}
\begin{figure}[!h]
    \centering
    \includegraphics[width=1.0\columnwidth]{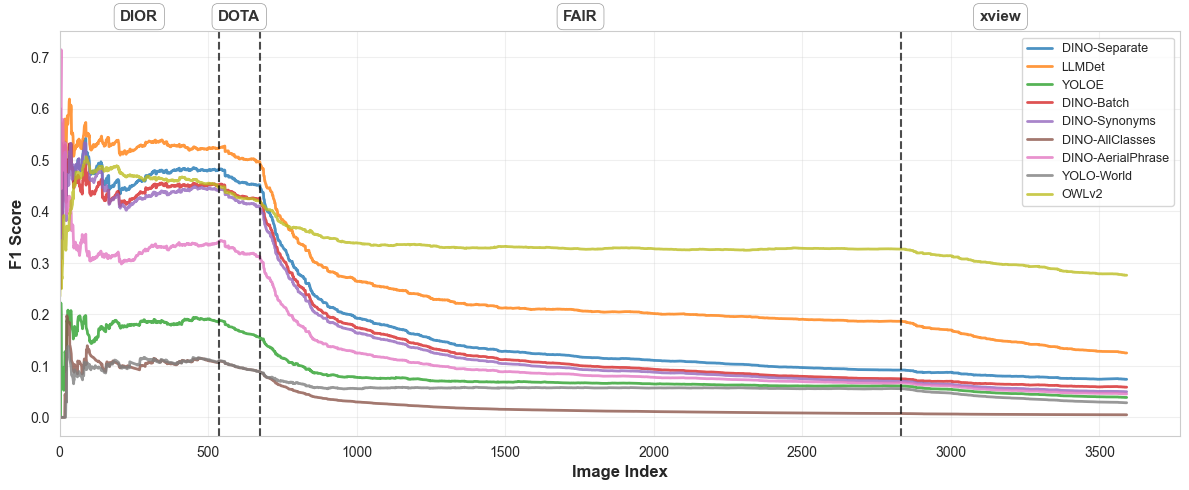}
    \caption{\textbf{F1-Score Evolution across LAE-80C.} The plot tracks the cumulative F1 score as the test set progresses through DIOR, DOTA, FAIR1M, and xView. Vertical dashed lines indicate dataset transitions. The sharp decline during the transitions to FAIR1M and xView highlight the models' inability to handle extreme scale variations.}
    \label{fig:dataset_evolution}
\end{figure}

As discussed earlier, LAE-80C is a collection of aerial images coming from the datasets DIOR, DOTA, FAIR, xView. We visualized the cumulative F1-score evolution, marking the transition between datasets with vertical dashed lines (Figure~\ref{fig:dataset_evolution}).

The evolution curve reveals a clear degradation pattern corresponding to the increasing difficulty of each sub-dataset:

\begin{enumerate}
    \item \textbf{DIOR (Region A):} Performance is highest in the initial phase for all models, with LLMDET achieving F1=0.53. DIOR images are typically object-centric with moderate density, allowing models to stabilize on decent f1-scores.
    \item \textbf{DOTA (Region B):} A slight performance drop occurs as the evaluation shifts to DOTA. This dataset introduces arbitrary object orientations and higher density, challenging the horizontal bounding box assumptions of standard detectors.
    \item \textbf{FAIR1M (Region C):} The most severe degradation is observed in the FAIR dataset, in which the average number of objects increases very much. Additionally, FAIR1M emphasizes fine-grained categories (37 distinct types of vehicles, ships, and buildings) that require subtle visual discrimination which is precisely the capability most degraded by aerial domain shift.
    \item \textbf{xView (Region D):} Performance starts degrading again in xView, which is characterized by extremely small Ground Sample Distance (GSD) and tiny objects (often $<20$ pixels).
\end{enumerate}

This confirms that while models may appear effective on "clean" aerial benchmarks (like DIOR), they fail to generalize to the complex, high-density, and multi-scale environments found in operational satellite data (FAIR1M, xView).

\section{Discussion}

Our systematic evaluation of state-of-the-art open-vocabulary detectors on aerial imagery reveals a fundamental mismatch between current VLM-based approaches and the requirements of UAV deployment. Three key insights emerge from our analysis.

\textbf{The Domain Gap is Architectural, Not Lexical.}
Our ablation studies demonstrate that prompt engineering—whether through domain-specific prefixes ("Aerial view of") or synonym expansion—fails to bridge the performance gap. In fact, these strategies often degrade results by introducing semantic drift or over-constraining the matching criteria. This failure suggests that the problem lies not in how we describe objects linguistically, but in how visual encoders extract features. Models pretrained on ground-level imagery learn to recognize objects through side-profile features (car wheels, building facades) that are entirely occluded in nadir aerial views. Text prompts cannot compensate for this fundamental visual mismatch.

\textbf{Semantic Confusion Dominates Visual Localization.}
The 15× performance improvement when moving from Global (80 classes) to Oracle (3.2 classes average) mode reveals that Grounding DINO can extract relevant visual features from aerial images—it simply cannot map them to the correct category when faced with overlapping taxonomies. LAE-80C's hierarchical structure ("vehicle" vs "small car" vs "passenger car") forces models to perform fine-grained discrimination that their CLIP-based text encoders were never trained for. This suggests that future research should focus on intelligent vocabulary filtering that reduces semantic ambiguity based on scene context.

\textbf{Dataset Diversity Exposes Brittleness.}
The sharp performance drops at dataset transitions (DIOR → DOTA → FAIR1M → xView) demonstrate that zero-shot generalization does not reliably hold. Models achieve borderline-acceptable results on clean, object-centric imagery (DIOR: F1=0.53 for LLMDet) but collapse when confronted with operational conditions: dense clusters (FAIR1M: F1=0.12), extreme scale variation (xView), or arbitrary orientations (DOTA). This brittleness reveals important implications for deployment, as a model validated on one aerial dataset cannot be trusted to generalize to different imaging conditions without extensive testing.

\textbf{Implications for Real-World Deployment.}
The precision-recall trade-off we observe creates a dilemma for practitioners. Conservative models are not appropriate for coverage-critical applications. Aggressive models detect most objects but generate many false alarms, requiring human verification that negates the automation benefit. Our results suggest that applying off-the-shelf natural-image OVD models to aerial tasks is currently fundamentally inadequate.

\section{Conclusion and Future Work}

This work presents a comprehensive benchmark of zero-shot open-vocabulary object detection on aerial imagery. We evaluated five state-of-the-art models (Grounding DINO, OWLv2, YOLO-World, YOLO-E, LLMDet) on the LAE-80C dataset, introducing a systematic experimental protocol that isolates semantic confusion from visual localization failure through Global, Oracle, and Single-Category inference modes. Our key findings establish critical limitations of current approaches:

\begin{itemize}
    \item Severe domain gap: All models experience dramatic performance degradation (best F1=0.276), with no architecture achieving balanced precision-recall trade-offs suitable for autonomous deployment.
    \item Semantic confusion as primary bottleneck: A 15× performance improvement from vocabulary reduction (Global → Oracle) demonstrates that overlapping taxonomies severely limit performance.
    \item Prompt engineering failures: Domain-specific text prefixes and synonym expansion degrade rather than improve results, confirming that lexical adaptation cannot compensate for visual encoder mismatch, at least in non-fine-tuned models.
    \item Dataset-dependent brittleness: Sharp performance drops across imaging conditions (DIOR → FAIR1M → xView) reveal that zero-shot generalization fails when confronted with density variation and extreme scales.
\end{itemize}

These findings suggest that the "open vocabulary" promise of VLM-based detection does not transfer from natural to aerial domains without fundamental architectural changes or domain-adaptive pretraining.Several paths could address the limitations we identify:

\begin{itemize}
\item Hierarchical Vocabulary Management: Intelligent prompt filtering that selects semantically compatible subsets based on scene context (urban/rural, altitude, sensor type) could reduce the semantic confusion that cripples Global mode.

\item Hybrid Detection Pipelines: Combining fast, specialized closed-set detectors for common categories with OVD models for rare objects could balance coverage and precision requirements.

\item Test-Time Adaptation: Lightweight adaptation mechanisms that adjust visual features or text embeddings using a few labeled aerial examples could bridge the domain gap without full retraining.

\item Alternative Evaluation Protocols: Moving beyond strict bounding box metrics to task-specific evaluation (e.g., "detect at least one damaged building in the scene") could better reflect real-world UAV deployment requirements.
\end{itemize}

Our benchmark and findings provide a foundation for these future efforts, establishing baseline performance expectations and highlighting the specific failure modes that aerial-specific OVD methods must address.

\bibliographystyle{IEEEtran}
\bibliography{refs}

\end{document}